\documentclass[letterpaper, 10 pt, conference]{ieeeconf}  % Comment this line out if you need a4paper

\IEEEoverridecommandlockouts                              % This command is only needed if 
\usepackage{graphicx}
\usepackage{adjustbox}
\usepackage{array}
\usepackage{bbding}
\usepackage{times,graphicx,amsmath,amssymb,booktabs,tabulary,multirow,overpic,xcolor}
\definecolor{color4}{rgb}{0.94,0.94,1}
\usepackage{colortbl}
\usepackage{makecell}
\usepackage{float}
\usepackage{url}
\title{\LARGE \bf
Learning Spectral Diffusion Prior for Hyperspectral Image Reconstruction
}

\author{Mingyang Yu, Zhijian Wu and Dingjiang Huang$^{*}$% <-this % stops a space
\thanks{*This work was supported by the National Natural Science Foundation of China under Grant 62072185, U1711262.}% <-this % stops a space
\thanks{Mingyang Yu, Zhijian Wu and Dingjiang Huang are with the School of Data Science and Engineering, East China Normal University,
        {\tt\small \{yumingyang,zjwu\_97\}@stu.ecnu.edu.cn, djhuang@dase.ecnu.edu.cn}}%
\thanks{Mingyang Yu and Zhijian Wu contributed equally to this work.}%
}

\begin{document}

\maketitle
\thispagestyle{empty}
\pagestyle{empty}

%%%%%%%%%%%%%%%%%%%%%%%%%%%%%%%%%%%%%%%%%%%%%%%%%%%%%%%%%%%%%%%%%%%%%%%%%%%%%%%%
\begin{abstract}
Hyperspectral image (HSI) reconstruction aims to recover 3D HSI from its degraded 2D measurements. Recently great progress has been made in deep learning-based methods, however, these methods often struggle to accurately capture high-frequency details of the HSI. To address this issue, this paper proposes a Spectral Diffusion Prior (SDP) that is implicitly learned from hyperspectral images using a diffusion model. Leveraging the powerful ability of the diffusion model to reconstruct details, this learned prior can significantly improve the performance when injected into the HSI model. To further improve the effectiveness of the learned prior, we also propose the Spectral Prior Injector Module (SPIM) to dynamically guide the model to recover the HSI details. We evaluate our method on two representative HSI methods: MST and BISRNet. Experimental results show that our method outperforms existing networks by about 0.5 dB, effectively improving the performance of HSI reconstruction.
\end{abstract}

%%%%%%%%%%%%%%%%%%%%%%%%%%%%%%%%%%%%%%%%%%%%%%%%%%%%%%%%%%%%%%%%%%%%%%%%%%%%%%%%
\section{Introduction}
\label{sec:intro}
Hyperspectral Image (HSI) is a special form of imagery that incorporates many channels beyond the regular RGB images. Each channel in HSI captures scenes at different wavelengths in the real world. Due to its ability to store more comprehensive information, HSI is frequently employed in various computer vision tasks, such as medical image processing~\cite{backman2000detection,meng2020snapshot}, object tracking~\cite{pan2003face,kim20123d}, and remote sensing~\cite{melgani2004classification,borengasser2007hyperspectral}, among others. However, traditional methods for capturing HSI have encountered certain challenges, including requiring a long time. Consequently, the Coded Aperture Snapshot Spectral Imaging (CASSI) system~\cite{wagadarikar2008single} was developed. This system captures images in a two-dimensional measurement~\cite{Yuan_review}, which offers greater convenience compared to conventional methods. However, a subsequent step is required to reconstruct the 2D measurement into a 3D HSI. There are currently several algorithms that utilize the 2D measurement provided by the CASSI system to reconstruct 3D HSI. The main approaches include model-based methods~\cite{bioucas2007new}, end-to-end methods~\cite{cai2022mask,cst}, and deep unfolding methods~\cite{cai2022degradation}. However, these algorithms face a common problem, which is the loss of high-frequency information~\cite{wu2023sfhn}. In particular, HSI typically contains more detailed information with richer high-frequency components than conventional RGB images. 

\begin{figure}[t]
	\begin{center}
		\begin{tabular}[t]{c} \hspace{-3.8mm} 
			\includegraphics[width=0.4\textwidth]{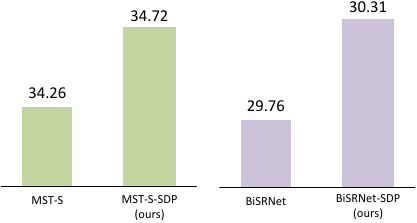}
		\end{tabular}
	\end{center}
	
	\caption{\small Illustrations of the comparison of the peak signal-to-noise ratio (PSNR) values with and without the integration of the SDP in the MST-S and BiSRNet models. In MST-S, the inclusion of SDP resulted in an improvement of 0.46 dB. Similarly, in BiSRNet, the inclusion of SDP led to an improvement of 0.55 dB. }
	\label{fig:1}
	
\end{figure}

\begin{figure*}[t]
	\begin{center}
		\begin{tabular}[t]{c} \hspace{-3mm}
			\includegraphics[width=1.0\textwidth]{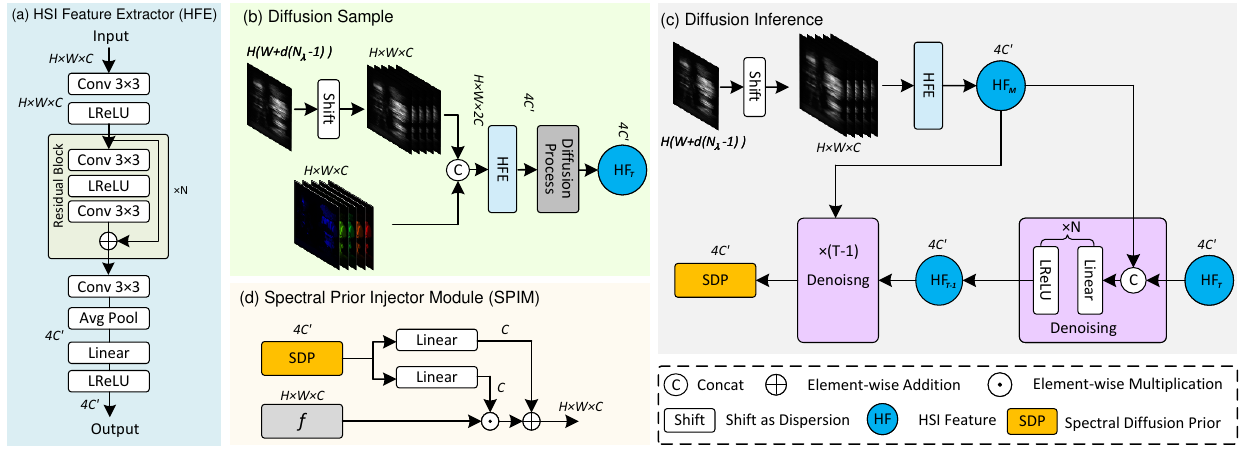}
		\end{tabular}
	\end{center}
	
	\caption{\small The overall architecture of the proposed method. (a) Illustration of the HSI feature extractor (HFE). (b) The sampling process of the spectral diffusion. (c) The training process of spectral diffusion. (d) Illustration of the spectral prior injector module (SPIM).}
	\label{fig:architecture}
	
\end{figure*}
Recently, Diffusion Models (DMs) \cite{ho2020denoising}, have achieved impressive results in density estimation and sample quality \cite{wang2022zero,xia2023diffir}. In contrast to other generative models, DMs utilize parameterized Markov chains to efficiently map high-quality samples from randomly sampled Gaussian noise to complex target distributions, allowing them to generate highly accurate target outputs. However, it is not feasible to apply DMs directly to the HSI reconstruction. On the one hand, DMs suffer from high computational complexity due to the multi-step sampling and inference process. On the other hand, generative models like DMs are prone to produce undesired artifacts when directly predicting the targeted HSI.

In this paper, we introduce a novel Spectral Diffusion Prior (SDP) that can be learned implicitly through diffusion models, which has the ability to reveal spectral detail information in HSI. By incorporating this information into the training methods, the performance of HSI reconstruction models can be significantly improved. To ensure that the diffusion model (DM) does not generate irrelevant information, we propose an HSI Feature Extractor (HFE) that extracts a compact feature from the 2D measurement. This extracted feature serves as guidance information for the DM. Subsequently, we train a spectral diffusion model to estimate a spectral prior from the extracted features. The diffusion process occurs in a low-dimensional feature space, which greatly reduces the model computation.
We further present a Spectral Prior Injection Module (SPIM) to guide the model in reconstructing high-quality HSI details from the spectral prior. Our SDP essentially follows a plug-and-play paradigm, making it easily incorporated into any existing network. We evaluated our SDP on two representative HSI reconstruction algorithms: MST and BISRNet. As shown in Fig \ref{fig:1}, our SDP achieves significant improvements of about 0.5 dB PSNR compared to the baseline models. These results demonstrate the effectiveness of our proposed method.

Our contributions can be summarized as follows:
\begin{itemize}
  \item In this paper, we explore the effect of diffusion prior to HSI reconstruction. Our method represents a significant step forward in the area of HSI reconstruction.
  \item We propose that a novel Spectral Diffusion Prior (SDP) that can be learned implicitly using a diffusion model. In addition, we propose a Spectral Prior Injection Module (SPIM) for guiding the model to learn HSI details from the spectral prior.
  \item We evaluate our method on MST and BISRNet. The experimental results show that our method can significantly improve the performance over the baselines.
\end{itemize}

\section{Related Work}

\subsection{High Spectral Image (HSI) Reconstruction Algorithm}

The methods for HSI reconstruction can be broadly classified into two categories: model-based approaches and learning-based approaches. Initially, model-based methods relied on hand-crafted image priors, such as low-rank, sparsity, and total variation, to formulate an optimization problem and then iteratively solve it. However, these methods were limited by low efficiency and heavy reliance on the empirical experience of hand-designed priors.

With the rapid development of deep learning in various fields, including image reconstruction, learning-based HSI reconstruction methods have emerged. These methods directly learn the mapping between the measurement and the HSI image, which has significantly improved the performance and speed of HSI reconstruction. Yet, these methods still lack interpretability.

Recent studies have proposed plug-and-play (PnP) algorithms and deep unfolding methods to address the limitations of previous approaches. These approaches combine model-based and learning-based techniques, where deep learning models replace hand-designed priors. The deep unfolding method enhances the quality of each reconstruction step, greatly reducing the number of iterations and improving the reconstruction speed.

\subsection{Diffusion Model}

Diffusion Models (DMs) have gained tremendous popularity in the field of generative models, primarily due to their exceptional ability to generate fine details. The core idea behind DMs is to parameterize the generative process as a Markov chain. The input is progressively corrupted with noise, and the model is then trained to predict the noise at each time step, effectively removing the noise and reconstructing the original signal.

This unique structure of DMs allows them to capture the coarse structure of the target in the early time steps and gradually refine the details in subsequent steps. This capability has made DMs a powerful tool for various computer vision tasks, including image reconstruction. However, the limited information capacity of one-dimensional vectors like SDP has hindered their performance when combined with reconstruction models that inherently possess the ability to capture spatial information.

\section{Method}
\vspace{-0.1mm}
\subsection{Overall Architecture}

The overall architecture of our method is illustrated in Fig \ref{fig:architecture}. Specifically, we employ the HSI Feature Extractor (HFE) to extract compact features from the 2D measurement. These extracted features serve as guidance information for a spectral diffusion model, which is tasked with generating Spectral Diffusion Priors (SDPs). Subsequently, a Spectral Prior Injection Module (SPIM) incorporates these SDPs into the input features. This integration guides the model to produce high-quality outputs by leveraging the spectral information contained within the SDPs. The model training process is divided into two stages. In the first stage, the HFE and spectral diffusion model are jointly trained to learn effective feature representations and generate meaningful SDPs. In the second stage, the entire architecture, including the SPIM, is fine-tuned to optimize the reconstruction performance. This two-stage training approach ensures that the model can effectively utilize the SDPs to enhance the quality of the reconstructed HSI.

\vspace{-2mm}
\subsection{Stage I: Base Model}
\vspace{-2mm}
In the first stage, we utilize the 2D measure and the corresponding 3D HSI to train a base model. We first shift back the measurement to get the initialized signal cube $\mathbf{H} \in \mathbb{R}^{H\times W \times N_{\lambda}}$, 
\vspace{-0.5mm}
\begin{equation}
\mathbf{H}(x,y,n_\lambda) = \mathbf{Y}(x,y-d(\lambda_n-\lambda_c)),
\vspace{-0.5mm}
\end{equation} 
Subsequently, the initial feature $\mathbf{H} \in \mathbb{R}^{H\times W \times N_{\lambda}}$ and the corresponding HSI $\mathbf{G} \in \mathbb{R}^{H\times W \times N_{\lambda}}$ are concatenated and are fed into the HFE to extract the HSI feature (HF).
\begin{equation}
\mathbf{HF}(x,y,2n_\lambda) =\text{HFE}(\text{Concat}(H,G))
\vspace{-0.5mm}
\end{equation} 
where $\mathbf{HF}  \in \mathbb{R}^{4C'}$ is a vector. $\text{HFE}$ consists of a series of ResBlocks\cite{he2016deep}, followed by a pooling and convolution operation. Our design has multiple advantages. Firstly, the extracted HF contains the high-frequency details of the HSIs while maintaining a reasonable correlation with the original measurements.
Secondly, the resulting HF is independent of spatial resolution. This gives our method great flexibility in processing inputs with varying sizes. Thirdly, the compact representation of HF can greatly reduce the difficulty of training subsequent diffusion models.  By providing a compact and informative representation of the high-frequency details, our method can make the training process more efficient and effective. This is beneficial for improving the overall performance of the HSI reconstruction system.

\subsection{Stage II: Spectral Diffusion Model}

In the second stage, we employ HFE to extract features from the measurement $\mathbf{H} \in \mathbb{R}^{H\times W \times N_{\lambda}}$ to obtain HSI feature $HF_M$.
\begin{equation}
\mathbf{HF_M} =\mathop{\text{HFE}}(H),
\end{equation} 
We generate a spectral diffusion prior by training a spectral diffusion model (SDM). The SDM transforms the HSI feature $x_0$ to the Gaussian noise $x_T \sim \mathcal{N}(0, 1)$ by $T$ iterations, which can be denoted as:
\begin{equation}
    q(\mathbf{x_t} | \mathbf{x_{t-1}}) = \mathcal{N}\left(\mathbf{x_t}; \sqrt{(1 - \beta_t)}\mathbf{x_{t-1}}, \beta_t \mathbf{I}\right)
\vspace{-0.5mm}
\label{eq:1}
\end{equation} 
As shown in Eq.\ref{eq:1}, $x_t$ is the noised image at time-step t, $\beta_t$ is the pre-set scale factor, and $\mathcal{N}$ represents the Gaussian distribution. This equation can be simplified as:
\begin{equation}
q(\mathbf{x_t} | \mathbf{x_{0}}) = \mathcal{N}\left(\mathbf{x_t}; \sqrt{\bar{\alpha_t}}\mathbf{x_{0}}, (1-\bar{\alpha})\mathbf{I}\right)
\label{eq:2}
\end{equation} 
where $\alpha_t=1-\beta_t,\quad \bar{\alpha}_t=\prod_{i=0}^t \alpha_i$.

\begin{table*}[t]
	\centering
 \caption{\small Quantitative results on 10 scenes in simulation. PSNR and SSIM are reported. Our SDP can significantly improve the performance of the base models.}\label{Tab:compare}
	\resizebox{0.75\textwidth}{!}{
		\setlength{\tabcolsep}{0.7mm}
		\centering
		\begin{tabular}{c|cccc|ccccc} 
			\toprule[0.15em]
			&  \multicolumn{2}{c}{ MST-S} & \multicolumn{2}{c|}{\bf MST-S-SDP(Ours)} & \multicolumn{2}{c}{BiSRNet} & \multicolumn{2}{c}{\bf BiSRNet-SDP(Ours)}\\
			Scene~&~PSNR~ & SSIM & \colorbox{color4}{PSNR}& \colorbox{color4}{SSIM}&~PSNR~ & SSIM & \colorbox{color4}{PSNR}  & \colorbox{color4}{SSIM}  \\
			\midrule[0.15em]
			1  & 34.71 & 0.930 & \colorbox{color4}{35.00\textcolor{red}{(+0.29)}} & \colorbox{color4}{0.938\textcolor{red}{(+0.008)}}& 30.95 & 0.847 & \colorbox{color4}{31.35\textcolor{red}{(+0.40)}} & \colorbox{color4}{0.870\textcolor{red}{(+0.023)}} \\
			2  & 34.45 & 0.925 & \colorbox{color4}{34.86\textcolor{red}{(+0.41)}} & \colorbox{color4}{0.934\textcolor{red}{(+0.009)}}& 29.21 & 0.791 & \colorbox{color4}{29.41\textcolor{red}{(+0.20)}} & \colorbox{color4}{0.811\textcolor{red}{(+0.020)}} \\
			3  & 35.32 & 0.943 & \colorbox{color4}{35.98\textcolor{red}{(+0.66)}} & \colorbox{color4}{0.949\textcolor{red}{(+0.006)}}& 29.11 & 0.828 & \colorbox{color4}{29.93\textcolor{red}{(+0.82)}} & \colorbox{color4}{0.856\textcolor{red}{(+0.028)}} \\
			4  & 41.50 & 0.967 & \colorbox{color4}{42.00\textcolor{red}{(+0.50)}} & \colorbox{color4}{0.973\textcolor{red}{(+0.006)}}& 35.91 & 0.903 & \colorbox{color4}{37.48\textcolor{red}{(+1.57)}} & \colorbox{color4}{0.924\textcolor{red}{(+0.021)}} \\
			5  & 31.90 & 0.933 & \colorbox{color4}{32.52\textcolor{red}{(+0.62)}} & \colorbox{color4}{0.941\textcolor{red}{(+0.008)}}& 28.19 & 0.827 & \colorbox{color4}{28.47\textcolor{red}{(+0.28)}} & \colorbox{color4}{0.853\textcolor{red}{(+0.026)}} \\
			6  & 33.85 & 0.943 & \colorbox{color4}{34.23\textcolor{red}{(+0.38)}} & \colorbox{color4}{0.950\textcolor{red}{(+0.007)}}& 30.22 & 0.863 & \colorbox{color4}{30.61\textcolor{red}{(+0.39)}} & \colorbox{color4}{0.879\textcolor{red}{(+0.016)}} \\
			7  & 32.69 & 0.911 & \colorbox{color4}{33.24\textcolor{red}{(+0.55)}} & \colorbox{color4}{0.920\textcolor{red}{(+0.009)}}& 27.85 & 0.800 & \colorbox{color4}{28.50\textcolor{red}{(+0.65)}} & \colorbox{color4}{0.822\textcolor{red}{(+0.022)}} \\
			8  & 31.69 & 0.933 & \colorbox{color4}{32.46\textcolor{red}{(+0.77)}} & \colorbox{color4}{0.946\textcolor{red}{(+0.013)}}& 28.82 & 0.843 & \colorbox{color4}{28.91\textcolor{red}{(+0.09)}} & \colorbox{color4}{0.856\textcolor{red}{(+0.013)}} \\
			9  & 34.67 & 0.939 & \colorbox{color4}{34.79\textcolor{red}{(+0.12)}} & \colorbox{color4}{0.940\textcolor{red}{(+0.001)}}& 29.46 & 0.832 & \colorbox{color4}{29.85\textcolor{red}{(+0.39)}} & \colorbox{color4}{0.860\textcolor{red}{(+0.028)}} \\
			10 & 31.82 & 0.926 & \colorbox{color4}{32.12\textcolor{red}{(+0.30)}} & \colorbox{color4}{0.936\textcolor{red}{(+0.010)}}& 27.88 & 0.800 & \colorbox{color4}{28.57\textcolor{red}{(+0.69)}} & \colorbox{color4}{0.828\textcolor{red}{(+0.028)}} \\
			\midrule
			Avg  & 34.26 & 0.935 & \colorbox{color4}{34.72\textcolor{red}{(+0.46)}} & \colorbox{color4}{0.943\textcolor{red}{(+0.008)}}& 29.76 & 0.833 & \colorbox{color4}{30.31\textcolor{red}{(+0.55)}} & \colorbox{color4}{0.856\textcolor{red}{(+0.023)}} \\
			\bottomrule[0.15em]
	\end{tabular}
 }
	
\end{table*}

In the reverse process, SDM samples a Gaussian random noise map $x_T$ and then denoise $x_T$ step by step until it reaches the HSI feature:
\begin{equation}
p(\mathbf{x_{t-1}} | \mathbf{x_t},\mathbf{x_{0}}) = \mathcal{N}\left(\mathbf{x_{t-1}}; \mu_t(\mathbf{x_t},\mathbf{x_0}), \sigma_t^2\mathbf{I}\right)
\label{eq:3}
\end{equation}
where mean $\boldsymbol{\mu}_{t}\left(\mathbf{x}_{t}, \mathbf{x}_{0}\right)=\frac{1}{\sqrt{\alpha_{t}}}\left(\mathbf{x}_{t}-\epsilon \frac{1-\alpha_{t}}{\sqrt{1-\bar{\alpha}_{t}}}\right)$ and variance $\sigma_{t}^{2}=\frac{1-\bar{\alpha}_{t-1}}{1-\bar{\alpha}_{t}} \beta_{t} .$ $\epsilon$ indicates the noise in $x_t$. SDM adopt a denoising network $\epsilon_{\theta}(x_t, t)$ to estimate the $\epsilon$. To get the HF, we train the $\epsilon_{\theta}(x_t, t)$ by optimizing the parameters $\theta$ of $\epsilon_{\theta}$ following:
\begin{equation}
\nabla_{\boldsymbol{\theta}}\left\|\epsilon-\epsilon_{\boldsymbol{\theta}}\left(\sqrt{\bar{\alpha}_{t}} \mathrm{x}_{0}+\epsilon \sqrt{1-\bar{\alpha}_{t}}, t\right)\right\|_{2}^{2} .
\label{eq:4}
\end{equation}
Since the diffusion process occurs only in a low-dimensional eigenspace, $\epsilon_{\theta}(x_t, t)$ does not need to be too long time step. In our design, we used a simple MLP structure to predict the noise and set the time step to 4. We use the features from $HF_M$ and SDM as reconstruction prior to predicting the resulting SDP.

\subsection{Spectral Prior Injector Module (SPIM)}
To effectively leverage the Spectral Diffusion Prior (SDP), we propose a Spectral Prior Injector Module (SPIM) designed to seamlessly integrate the learned prior into the model. Within the SPIM, we employ linear projections to transform the SDP into two distinct modulation vectors. These vectors serve as the means to modulate the input features, ensuring that the spectral information is effectively incorporated into the model's processing pipeline. Specifically, the spectral prior is injected into the input features through a combination of element-wise multiplication and addition operations between the input features and the modulation vectors. This approach allows the model to utilize the SDP in a way that enhances the reconstruction process by emphasizing important spectral details and improving the overall quality of the output. By integrating the SDP through these operations, the model gains a guided mechanism to focus on critical spectral information, leading to more accurate and detailed HSI reconstructions.
\begin{equation}
F^{'} =(W^1 \text{SDP}) \odot F + W^2 \text{SDP}+F
\end{equation} 
where $W^i$ denotes linear projection, $\odot$ is element-wise multiplication.

\section{Experiment}
\subsection{Experimental Settings}

\textbf{Datasets} Following previous works, we use two commonly used datasets, CAVE and KAIST, for training. The CAVE dataset consists of 32 hyperspectral images, each containing 31 spectral channels, with a spatial size of $512\times512$. The KAIST dataset comprises 30 hyperspectral images, also with 31 spectral channels, and a size of $2704\times3376$. Following the setup of TSA-Net, we used the same mask size of 256 and simulated the wavelength range of 450$\sim$650nm using spectral interpolation. We utilized the CAVE dataset for training and applied ten scenes from the KAIST dataset for testing purposes.

\begin{figure*}[t]
\centering
\includegraphics[width=0.85\textwidth]{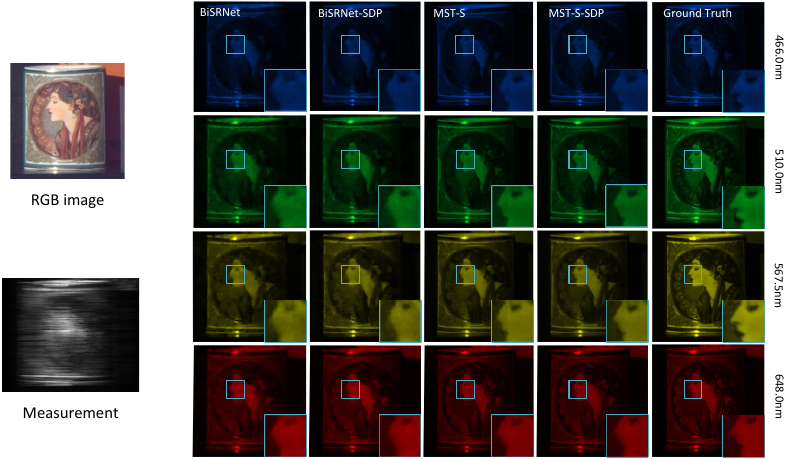}
\caption{\small Qualitative comparison of the four spectral bands in the hyperspectral image of Scene 1. Our SDP can facilitate the recovery of finer details. Please zoom in for a better view. }
\label{fig:compare}
\end{figure*}

\noindent

\textbf{Implementation Details} For the experiments on simulated data, spatial blocks of size $256\times256$ were cropped from the three-dimensional cubes and fed into the network. To reconstruct real hyperspectral images, the block size was set to $660\times660$ to match the measurements in the real world. The displacement step size d in dispersion was set to 2. Therefore, the measurement dimensions for the simulated and real HSI datasets were $256\times310$ and $660\times714$, respectively. Data augmentation techniques such as random flipping and rotation were applied.
To ensure fairness in testing the plugin, the training in Stage I followed the same settings as the base model, except for the learning rate. Due to the addition of the plugin, a large learning rate could lead to training instability. Therefore, the learning rate was set to half of the original model's learning rate. In Stage II training, a total of 50 epochs were conducted. The first five epochs focused solely on training the diffusion model with a learning rate of 0.0001. Adam optimizer was used with $\beta_1 = 0.9$ and $\beta_2 = 0.999$. The MSE loss function was employed for optimization. The evaluation metrics used were PSNR and SSIM.

\begin{table}[t!]\small
	\centering
 \caption{ Ablation study of the SPIM. }\label{Tab:Ablation}
 % \vspace{2mm}

		\centering
		\begin{tabular}{c|c|c|c}
			\toprule
			  Element-wise Multiplication&Addition&  PSNR& SSIM \\
			\midrule
   \XSolidBrush&\XSolidBrush&29.76&0.833\\
			 \Checkmark&\XSolidBrush&29.49 & 0.820 \\
                  \XSolidBrush&\Checkmark&30.00&0.839\\
                 \Checkmark&\Checkmark&\textbf{30.31}&\textbf{0.856} \\
			\bottomrule
	\end{tabular}
 
\end{table}

\begin{table*}[t!]
\begin{center}
\setlength{\abovecaptionskip}{0.cm}
\caption{Quantitative results on 10 scenes in simulation. PSNR and SSIM are reported.} \label{tab:cap2}
\resizebox{\textwidth}{!}{
\begin{tabular}{l|r|c|c|c|c|c|c|c|c|c|c|c}
  \toprule
  % after \\: \hline or \cline{col1-col2} \cline{col3-col4} ...
  Methods & Params & s1 & s2 & s3 & s4 & s5 & s6 & s7 & s8 & s9 & s10 & avg
  \\
  \midrule
  TwIST\cite{bioucas2007new} & - & 25.16/0.700 & 23.02/0.604 & 21.40/0.711 & 30.19/0.851 & 21.41/0.635 & 20.95/0.644 & 22.20/0.643 & 21.82/0.650 & 22.42/0.690 &  22.67/0.569 & 23.12/0.669\\
  GAP-TV\cite{yuan2016generalized}  & - & 26.82/0.754 & 22.89/0.610 & 26.31/0.802 & 30.65/0.852 & 23.64/0.703 & 21.85/0.663 & 23.76/0.688 & 21.98/0.655 & 22.63/0.682 & 23.10/0.584 & 24.36/0.669 \\
  DeSCI\cite{liu2018rank} & - & 27.13/0.748 & 23.04/0.620 & 26.62/0.818 & 34.96/0.897 & 23.94/0.706 & 22.38/0.683 & 24.45/0.743 & 22.03/0.673 & 24.56/0.732 & 23.59/0.587 & 25.27/0.721 \\
  $\lambda$-Net\cite{miao2019net} & 62.64M & 30.10/0.849 & 28.49/0.805 & 27.73/0.870 & 37.01/0.934 & 26.19/0.817 & 28.64/0.853 & 26.47/0.806 & 26.09/0.831 & 27.50/0.826 & 27.13/0.816 & 28.53/0.841 \\
  ADMMNet\cite{ma2019deep} & 4.27M & 34.12/0.918 & 33.62/0.902 & 35.04/0.931 & 41.15/0.966 & 31.82/0.922 & 32.54/0.924 & 32.42/0.896 & 30.74/0.907 & 33.75/0.915 & 30.68/0.895 & 33.58/0.918 \\
  TSA-Net\cite{meng2020end}  & 44.25M & 32.03/0.892 & 31.00/0.858 & 32.25/0.915 & 39.19/0.953 & 29.39/0.884 & 31.44/0.908 & 30.32/0.878 & 29.35/0.888 & 30.01/0.890 & 29.59/0.874 & 31.46/0.894 \\
  DGSMP\cite{huang2021deep}  & 3.76M & 33.26/0.915 & 32.09/0.898 & 33.06/0.925 & 40.54/0.964 & 28.86/0.882 & 33.08/0.937 & 30.74/0.886 & 31.55/0.923 & 31.66/0.911 & 31.44/0.925 & 32.63/0.917 \\
  MST-S\cite{cai2022mask} & 0.93M & 34.71/0.930 & 34.45/0.925 & 35.32/0.943 & 41.50/0.967 & 31.90/0.933 & 33.85/0.943 & 32.69/0.911 & 31.69/0.933 & 34.67/0.939 & 31.82/0.926 & 34.26/0.935\\
  BiSRNet\cite{cai2024binarized} & 0.04M & 30.95/0.847 & 29.21/0.791 & 29.11/0.828 & 35.91/0.903 & 28.19/0.827 & 30.22/0.863 & 27.85/0.800 & 28.82/0.843 & 29.46/0.832 & 27.88/0.800 & 29.76/0.833\\

  \midrule
   \rowcolor[gray]{.85} \textbf{BiSRNet-SDP} & 0.06M & 31.35/0.870 & 29.41/0.810 & 29.93/0.856 & 37.48/0.924 & 28.47/0.853 & 30.61/0.879 & 28.50/0.822 & 28.91/0.856 & 29.85/0.860 & 28.57/0.828 & 30.31/0.856 \\
 \rowcolor[gray]{.85} \textbf{MST-S-SDP} & 1.30M & \textbf{35.00/0.938} & \textbf{34.86/0.934} & \textbf{35.98/0.949} & \textbf{42.00/0.973} & \textbf{32.52/0.941} & \textbf{34.23/0.950} & \textbf{33.24/0.920} & \textbf{32.46/0.946} & \textbf{34.79/0.940} & \textbf{32.12/0.936} & \textbf{34.72/0.943} \\

  \bottomrule
\end{tabular}
}
\end{center}
\end{table*}

\begin{figure*}[t]
	\begin{center}
		\begin{tabular}[t]{c} \hspace{-3mm}
			\includegraphics[width=1.0\textwidth]{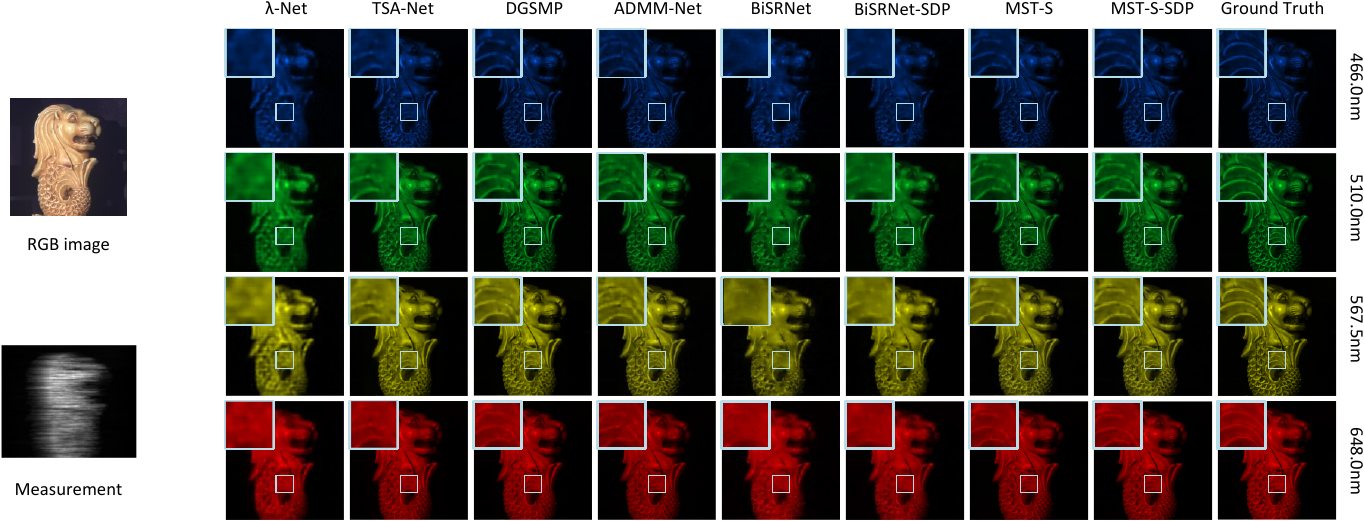}
		\end{tabular}
	\end{center}
	
	\caption{\small Comparison of MST-S-SDP and BiSRNet-SDP with other state-of-the-art models in the reconstruction of four spectral bands in Scene 10 hyperspectral image. Both MST-S-SDP and BiSRNet-SDP can recover structure and detail accurately. Please zoom in for a better view.}
	\label{fig:allcompare}
	
\end{figure*}

\subsection{Investigation of SDP}
To demonstrate the effectiveness of our proposed method, we conducted experiments on two representative models, namely MST-S and BiSRNet. The quantitative comparison results are presented in Table \ref{Tab:compare}. It can be observed that when our method is applied to MST-S, it achieves an improvement of 0.46dB in terms of PSNR. Similarly, when integrated with BiSRNet, our method leads to a PSNR enhancement of 0.55dB. These results clearly indicate that our approach can effectively boost the performance of existing HSI reconstruction models.
To further illustrate the benefits of our method, we also performed a qualitative comparison, as shown in Fig \ref{fig:compare}. From the figure, it is evident that our Spectral Diffusion Prior (SDP) significantly improves the model's ability to recover fine details. The reconstructed images produced by our method exhibit richer texture and more accurate structural information compared to those generated by the original models. This visual improvement underscores the practical advantages of incorporating our SDP into HSI reconstruction frameworks.
In summary, both the quantitative and qualitative results demonstrate the superiority of our method. By enhancing the models' capacity to capture and reconstruct spectral details, our approach offers a valuable advancement in the field of HSI reconstruction. These findings validate the effectiveness of our proposed method and highlight its potential for practical applications.

% We selected two models, MST-S and BiSRNet, to test our extension. These models were chosen because they represent different architectures and can demonstrate the significant improvement achieved by SDP across different model architectures. Detailed comparisons of scene details in the test dataset can be seen in Figure~\ref{fig:compare}, where it is evident that the results with the SDP plugin outperform the original models in terms of spatial high-frequency details.

\subsection{Ablation Study of SPIM}
We further analyze the impact of different components in the Spectral Prior Injector Module (SPIM) on the overall performance. To this end, we conduct experiments on the BiSRNet model, ensuring that all experimental settings remain consistent across the trials. The results of these experiments are presented in Table \ref{Tab:Ablation}. From the table, it is evident that relying solely on element-wise multiplication or addition yields suboptimal results. Notably, using only element-wise multiplication even leads to a degradation in performance. In contrast, our proposed SPIM leverages a combination of both element-wise multiplication and addition to achieve significant improvements.
We posit that this diverse aggregation strategy enables the model to receive different types of representations. By integrating both strategies, the model can more effectively utilize the learned spectral diffusion prior information. This dual approach allows for a more comprehensive and nuanced incorporation of the spectral priors, thereby enhancing the model's ability to reconstruct detailed and accurate spectral images. The results clearly indicate that the combination of both element-wise multiplication and addition within SPIM is crucial for optimizing the performance of the HSI reconstruction model.

\subsection{Comparisons with State-of-the-Arts}
We conducted a comprehensive comparison of our model with several other state-of-the-art models to evaluate its performance and effectiveness. The quantitative comparison results are presented in Table \ref{tab:cap2}. Despite our approach not being primarily aimed at outperforming state-of-the-art models, the results indicate that our model achieves the best performance among the compared models, even with a limited number of parameters. This suggests that our approach is not only efficient in terms of resource usage but also highly effective in HSI reconstruction.
Furthermore, to provide a more complete picture of our model's capabilities, we performed a qualitative comparison of the reconstruction results from different methods. The results of this comparison are illustrated in Fig \ref{fig:allcompare}. Visual inspection of the reconstructed images reveals that our method produces more visually promising results compared to competing algorithms. Specifically, other algorithms often fail to recover sufficient structural and textural details, leading to less accurate and less visually appealing reconstructions. In contrast, our method demonstrates a superior ability to capture and reproduce these details, resulting in images that are closer to the ground truth.
These findings collectively indicate that our method is more powerful and effective in recovering structural content and texture details in HSI reconstruction. This can be attributed to the innovative design of our model, which efficiently leverages the learned spectral diffusion prior information to guide the reconstruction process. The results of both the quantitative and qualitative comparisons serve as strong evidence of the effectiveness and efficiency of our proposed approach.

\section{Conclusion}
In this paper, we propose a novel approach termed Spectral Diffusion Prior (SDP), which utilizes a diffusion model to implicitly learn from 2D measurements and uncover spectral detail information. To enhance the reconstruction performance, we introduce the Spectral Prior Injector Module (SPIM) that employs a gating mechanism to regulate the learned SDPs and steer the model to focus on detailed information. We evaluate the proposed SDP in a plug-and-play framework on two representative HSI reconstruction networks. The experimental results illustrate the effectiveness of our approach. In the future, we plan to explore additional architectures to further learn and extend SDP.

\bibliographystyle{IEEEbib}
\bibliography{main}

\begin{thebibliography}{10}

\bibitem{backman2000detection}
V~Backman, Michael~B Wallace, LT~Perelman, JT~Arendt, R~Gurjar, MG~M{\"u}ller, Q~Zhang, G~Zonios, E~Kline, T~McGillican, et~al.,
\newblock ``Detection of preinvasive cancer cells,''
\newblock {\em Nature}, vol. 406, no. 6791, pp. 35--36, 2000.

\bibitem{meng2020snapshot}
Ziyi Meng, Mu~Qiao, Jiawei Ma, Zhenming Yu, Kun Xu, and Xin Yuan,
\newblock ``Snapshot multispectral endomicroscopy,''
\newblock {\em Optics Letters}, vol. 45, no. 14, pp. 3897--3900, 2020.

\bibitem{pan2003face}
Zhihong Pan, Glenn Healey, Manish Prasad, and Bruce Tromberg,
\newblock ``Face recognition in hyperspectral images,''
\newblock {\em IEEE TPAMI}, vol. 25, no. 12, pp. 1552--1560, 2003.

\bibitem{kim20123d}
Min~H Kim, Todd~Alan Harvey, David~S Kittle, Holly Rushmeier, Julie Dorsey, Richard~O Prum, and David~J Brady,
\newblock ``3d imaging spectroscopy for measuring hyperspectral patterns on solid objects,''
\newblock {\em ACM TOG}, vol. 31, no. 4, pp. 1--11, 2012.

\bibitem{melgani2004classification}
Farid Melgani and Lorenzo Bruzzone,
\newblock ``Classification of hyperspectral remote sensing images with support vector machines,''
\newblock {\em IEEE TGRS}, vol. 42, no. 8, pp. 1778--1790, 2004.

\bibitem{borengasser2007hyperspectral}
Marcus Borengasser, William~S Hungate, and Russell Watkins,
\newblock {\em Hyperspectral remote sensing: principles and applications},
\newblock CRC press, 2007.

\bibitem{wagadarikar2008single}
Ashwin Wagadarikar, Renu John, Rebecca Willett, and David Brady,
\newblock ``Single disperser design for coded aperture snapshot spectral imaging,''
\newblock vol. 47, no. 10, pp. B44--B51, 2008.

\bibitem{Yuan_review}
Xin Yuan, David~J Brady, and Aggelos~K Katsaggelos,
\newblock ``Snapshot compressive imaging: Theory, algorithms, and applications,''
\newblock {\em IEEE SPM}, 2021.

\bibitem{bioucas2007new}
Jos{\'e}~M Bioucas-Dias and M{\'a}rio~AT Figueiredo,
\newblock ``A new twist: Two-step iterative shrinkage/thresholding algorithms for image restoration,''
\newblock {\em IEEE TIP}, vol. 16, no. 12, pp. 2992--3004, 2007.

\bibitem{cai2022mask}
Yuanhao Cai, Jing Lin, Xiaowan Hu, Haoqian Wang, Xin Yuan, Yulun Zhang, Radu Timofte, and Luc Van~Gool,
\newblock ``Mask-guided spectral-wise transformer for efficient hyperspectral image reconstruction,''
\newblock in {\em CVPR}, 2022, pp. 17502--17511.

\bibitem{cst}
Yuanhao Cai, Jing Lin, Xiaowan Hu, Haoqian Wang, Xin Yuan, Yulun Zhang, Radu Timofte, and Luc~Van Gool,
\newblock ``Coarse-to-fine sparse transformer for hyperspectral image reconstruction,''
\newblock in {\em ECCV}, 2022.

\bibitem{cai2022degradation}
Yuanhao Cai, Jing Lin, Haoqian Wang, Xin Yuan, Henghui Ding, Yulun Zhang, Radu Timofte, and Luc~V Gool,
\newblock ``Degradation-aware unfolding half-shuffle transformer for spectral compressive imaging,''
\newblock {\em NeurIPS}, vol. 35, pp. 37749--37761, 2022.

\bibitem{wu2023sfhn}
Zhijian Wu, Wenhui Liu, Jun Li, Chang Xu, and Dingjiang Huang,
\newblock ``Sfhn: Spatial-frequency domain hybrid network for image super-resolution,''
\newblock {\em IEEE TCSVT}, 2023.

\bibitem{ho2020denoising}
Jonathan Ho, Ajay Jain, and Pieter Abbeel,
\newblock ``Denoising diffusion probabilistic models,''
\newblock {\em NeurIPS}, vol. 33, pp. 6840--6851, 2020.

\bibitem{wang2022zero}
Yinhuai Wang, Jiwen Yu, and Jian Zhang,
\newblock ``Zero-shot image restoration using denoising diffusion null-space model,''
\newblock {\em ICLR}, 2022.

\bibitem{xia2023diffir}
Bin Xia, Yulun Zhang, Shiyin Wang, Yitong Wang, Xinglong Wu, Yapeng Tian, Wenming Yang, and Luc Van~Gool,
\newblock ``Diffir: Efficient diffusion model for image restoration,''
\newblock in {\em ICCV}, 2023, pp. 13095--13105.

\bibitem{he2016deep}
Kaiming He, Xiangyu Zhang, Shaoqing Ren, and Jian Sun,
\newblock ``Deep residual learning for image recognition,''
\newblock in {\em CVPR}, 2016, pp. 770--778.

\bibitem{yuan2016generalized}
Xin Yuan,
\newblock ``Generalized alternating projection based total variation minimization for compressive sensing,''
\newblock in {\em ICIP}, 2016, pp. 2539--2543.

\bibitem{liu2018rank}
Yang Liu, Xin Yuan, Jinli Suo, David~J Brady, and Qionghai Dai,
\newblock ``Rank minimization for snapshot compressive imaging,''
\newblock {\em IEEE TPAMI}, vol. 41, no. 12, pp. 2990--3006, 2018.

\bibitem{miao2019net}
Xin Miao, Xin Yuan, Yunchen Pu, and Vassilis Athitsos,
\newblock ``l-net: Reconstruct hyperspectral images from a snapshot measurement,''
\newblock in {\em ICCV}, 2019, pp. 4059--4069.

\bibitem{ma2019deep}
Jiawei Ma, Xiao-Yang Liu, Zheng Shou, and Xin Yuan,
\newblock ``Deep tensor admm-net for snapshot compressive imaging,''
\newblock in {\em ICCV}, 2019, pp. 10223--10232.

\bibitem{meng2020end}
Ziyi Meng, Jiawei Ma, and Xin Yuan,
\newblock ``End-to-end low cost compressive spectral imaging with spatial-spectral self-attention,''
\newblock in {\em ECCV}. Springer, 2020, pp. 187--204.

\bibitem{huang2021deep}
Tao Huang, Weisheng Dong, Xin Yuan, Jinjian Wu, and Guangming Shi,
\newblock ``Deep gaussian scale mixture prior for spectral compressive imaging,''
\newblock in {\em CVPR}, 2021, pp. 16216--16225.

\bibitem{cai2024binarized}
Yuanhao Cai, Yuxin Zheng, Jing Lin, Xin Yuan, Yulun Zhang, and Haoqian Wang,
\newblock ``Binarized spectral compressive imaging,''
\newblock {\em NeurIPS}, vol. 36, 2024.

\end{thebibliography}

\end{document}